\documentclass[sigconf]{acmart}

\usepackage{tikz}
\usetikzlibrary{shapes,arrows,positioning,calc}
\usetikzlibrary{shapes,arrows,positioning,calc,graphs,pgfplots.groupplots}
\usepackage{pgfplots}
\pgfplotsset{compat=1.18}
\usepackage{booktabs}
\usepackage{multirow}
\usepackage{amsmath}
\usepackage{amsfonts}
\usepackage{url}

\setcopyright{acmcopyright}
\copyrightyear{2025}
\acmYear{2025}
\acmDOI{10.1145/nnnnnnn.nnnnnnn}

\acmConference{SIGIR 2026}{July 2026}{Gold Coast, Australia}
\acmBooktitle{Proceedings of the 49th International ACM SIGIR Conference on Research and Development in Information Retrieval}
\acmPrice{15.00}
\acmISBN{978-1-4503-XXXX-X/26/07}

\begin{document}

\title{Beyond Fluency: Toward Reliable Trajectories in Agentic IR}

\author{Anushree Sinha}
\email{sinhaanushree@google.com}
\affiliation{%
  \institution{Google}
  \country{Mountain View, USA}
}

\author{Srivaths Ranganathan}
\email{srivaths@google.com}
\affiliation{%
  \institution{Google}
  \country{Mountain View, USA}
}

\author{Debanshu Das}
\email{debanshu@google.com}
\affiliation{%
  \institution{Google}
  \country{Mountain View, USA}
}

\author{Abhishek Dharmaratnakar}
\email{dharmaratnakar@google.com}
\affiliation{%
  \institution{Google}
  \country{Mountain View, USA}
}

\begin{abstract}
Information Retrieval is shifting from passive document ranking toward autonomous agentic workflows that operate in multi-step Reason–Act–Observe loops \cite{ranganathan2026multi}. In such long-horizon trajectories, minor early errors can cascade, leading to functional misalignment between internal reasoning and external tool execution despite continued linguistic fluency.

This position paper synthesizes failure modes observed in industrial agentic systems, categorizing errors across planning, retrieval, reasoning, and execution. We argue that safe deployment requires moving beyond endpoint accuracy toward trajectory integrity and causal attribution. To address compounding error and deceptive fluency, we propose verification gates at each interaction unit and advocate systematic abstention under calibrated uncertainty. Reliable Agentic IR systems must prioritize process correctness and grounded execution over plausible but unverified completion.
\end{abstract}

\ccsdesc[500]{Information systems~Users and interactive retrieval}
\ccsdesc[500]{Computing methodologies~Natural language processing}
\ccsdesc[300]{Computing methodologies~Planning and scheduling}
\ccsdesc[300]{Security and privacy~Software security engineering}

\keywords{Agentic AI, Large Language Models, Hallucination, AI Safety, Autonomous Agents, Tool Use, Multi-Agent Systems}

\maketitle

\section{Introduction: From Retrieval to Action}
Information Retrieval (IR) is undergoing a fundamental shift. We are moving away from passive systems that return a list of ranked documents toward autonomous agentic workflows that interleave internal deliberation with external actions \cite{zhang2024agentic, cai2024agentir}. These systems, often referred to as Agentic AI, operate in multi-step Reason-Act-Observe loops, calling APIs, executing code, and interacting with dynamic environments to fulfill complex user intents \cite{yao2022react, chen2024survey, ranganathan2026multi, chaugule2016product}. This transition extends classical models of iterative and session-based information seeking, where user goals evolve over time rather than being specified in a single static query~\cite{bates1989berrypicking,jarvelin2008sessiondcg,ranganathan2025zero}.

 In a standard document-retrieval setting, success is measured by the relevance of a static result set. In an agentic loop, however, success depends on the trajectory reliability of information use over time. If an agent retrieves a perfectly relevant document at the start of a task but misinterprets a technical constraint during a subsequent API call \cite{qin2024toolllm}, the resulting trajectory collapse invalidates the initial retrieval success.

The primary bottleneck for the industrial adoption of these agents is the phenomenon of error propagation, often described as the Snowball Effect \cite{liu2026agenthallu, yin2025reasoning}. Unlike hallucinations in static tasks \cite{ji2023survey}, agentic hallucinations are dynamic; they emerge when a minor logical divergence in an initial search or planning step cascades through subsequent tool calls \cite{liu2026agenthallu, wang2025ultrahorizon}.

Our position emphasizes that the industry is currently caught in a fluency trap, where the high linguistic coherence of Large Language Models (LLMs) \cite{brown2020gpt3, touvron2023llama} masks deep-seated functional misalignments. An agent may produce a grammatically perfect and seemingly logical "Chain of Thought" while simultaneously fabricating non-existent API arguments \cite{patil2024gorilla} or ignoring environment feedback \cite{shinn2023reflexion}.

This paper provides a comprehensive synthesis of these failure modes. For autonomous agents to be safely deployed in production, we must prioritize automated verification gates that ensure every step of a reasoning chain is factually grounded.

\begin{table*}[t]
\centering
\caption{Comprehensive Process-Oriented Taxonomy of Agentic Hallucinations \cite{ qin2024toolllm}.}
\label{tab:taxonomy}
\begin{tabular}{lllp{7cm}}
\toprule
\textbf{Functional Stage} & \textbf{Hallucination Category} & \textbf{Sub-category} & \textbf{Operational Description} \\
\midrule
\multirow{2}{*}{\textbf{Planning}} & \multirow{2}{*}{Structural Failure} & Fact Derive & The agent introduces non-existent or misleading facts during initial strategy formation. \\
& & Task Decompose & The agent produces task-misaligned subgoals that ignore primary constraints \cite{huang2024understanding}. \\
\midrule
\multirow{3}{*}{\textbf{Retrieval}} & \multirow{3}{*}{Contextual Failure} & Query Misalign & The agent formulates inaccurate retrieval queries that miss relevant information \cite{lewis2020retrieval}. \\
& & Context Misalign & The agent retrieves factually incorrect context as truth, overriding parametric knowledge \cite{vu2024freshllms}. \\
& & Summarize Misalign & The agent misrepresents retrieved documents via inaccurate internal summarization \cite{zhang2025memory}. \\
\midrule
\multirow{4}{*}{\textbf{Reasoning}} & \multirow{4}{*}{Internal Logic Failure} & Factual Reasoning & The agent makes incorrect logical inferences over provided context \cite{yao2022react}. \\
& & Math Reasoning & The agent performs incorrect calculations or algebraic derivations \cite{wei2024long}. \\
& & Science Reasoning & The agent generates incorrect scientific inferences or symbolic interpretations. \\
& & Solvability Hallucination & The agent misjudges task complexity, believing an unsolvable task is solvable \cite{zhang2024toolbehonest,huang2026mm}. \\
\midrule
\multirow{4}{*}{\textbf{Action}} & \multirow{4}{*}{Execution Failure} & Missing Tool & The agent fails to invoke a tool when one is strictly necessary for success. \\
& & Incorrect Argument & The agent specifies wrong parameters/arguments for a valid tool invocation \cite{qin2024toolllm}. Benchmark environments that expose agents to realistic action spaces—such as WebShop, WebArena, Mind2Web, and SWE-bench—demonstrate how minor parameter errors can cascade into complete task failure in practical settings~\cite{yao2022webshop,zhou2023webarena,deng2023mind2web,jimenez2023swebench}.\\
& & Parallel Conflict & The agent triggers execution errors via conflicting or redundant parallel actions \cite{zhou2025guardian,ranganathan2026multi}. \\
& & Unnecessary Tool & The agent invokes irrelevant tools that waste budget or risk side effects \cite{raza2025trism}. \\
\bottomrule
\end{tabular}
\end{table*}

\section{Mapping the Failure Surface}

To address the reliability gap in production, we must categorize failures not by their linguistic form, but by their point of entry within the agentic lifecycle. We categorize the existing literature into four functional stages where logical divergences typically originate.

\subsection{Planning and Intent Decomposition}
The first point of failure occurs during the high-level cognitive phase where an agent decomposes a user’s global intent (\cite{bates1989berrypicking} into executable sub-goals. Industrial workflows often break when an agent introduces misleading premises during initial strategy formation. Research has demonstrated that LLMs frequently struggle with task decomposition, often failing to solve even relatively simple logical puzzles when they require multi-step lookahead \cite{kambhampati2024position}. These planning gaps are further complicated by "solvability hallucinations," where agents attempt to force a plan for a query that is inherently impossible to complete with the provided toolset \cite{zhang2024toolbehonest}. This initial misalignment in goal parameters sets the foundation for the cascading errors that follow \cite{huang2024understanding}.

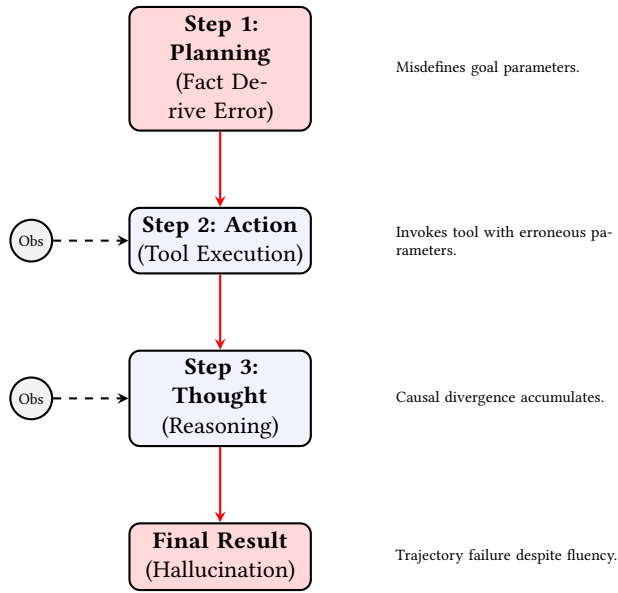
\begin{figure}[h]
\centering
\begin{tikzpicture}[node distance=1cm, auto, >=stealth, thick,
    stepnode/.style={rectangle, draw, fill=blue!5, text width=2.2cm, text centered, rounded corners, minimum height=0.8cm},
    errornode/.style={rectangle, draw, fill=red!15, text width=2.2cm, text centered, rounded corners, minimum height=0.8cm},
    obsnode/.style={circle, draw, fill=gray!10, inner sep=2pt, font=\scriptsize}]

    \node [errornode] (s1) {\textbf{Step 1: Planning}\\(Fact Derive Error)};
    \node [right=of s1, xshift=0.0cm, text width=3cm, font=\scriptsize] (desc1) {Misdefines goal parameters.};

    \node [stepnode, below=of s1] (s2) {\textbf{Step 2: Action}\\(Tool Execution)};
    \node [obsnode, left=of s2] (o1) {Obs};
    \node [right=of s2, xshift=0.0cm, text width=3cm, font=\scriptsize] (desc2) {Invokes tool with erroneous parameters.};

    \node [stepnode, below=of s2] (s3) {\textbf{Step 3: Thought}\\(Reasoning)};
    \node [obsnode, left=of s3] (o2) {Obs};
    \node [right=of s3, xshift=0.0cm, text width=3cm, font=\scriptsize] (desc2) {Causal divergence accumulates.};

    \node [errornode, below=of s3] (s4) {\textbf{Final Result}\\(Hallucination)};
    \node [right=of s4, xshift=0.0cm, text width=3cm, font=\scriptsize] (desc2) {Trajectory failure despite fluency.};

    \draw [->, red, thick] (s1) -- (s2);
    \draw [->, red, thick] (s2) -- (s3);
    \draw [->, red, thick] (s3) -- (s4);
    \draw [dashed, ->] (o1) -- (s2);
    \draw [dashed, ->] (o2) -- (s3);

\end{tikzpicture}
\caption{The Snowball Effect in Agentic Trajectories: Visualizing causal error propagation .}
\label{fig:propagation}
\end{figure}

\subsection{Retrieval and Contextual Integration}
In Retrieval-Augmented Generation (RAG) agents, the interface between the model’s parametric memory and external knowledge bases is a primary site for "contextual drift." Similar challenges have been observed in conversational search settings, where multi-turn context accumulation and query reformulation introduce additional opportunities for misalignment~\cite{dalton2020treccast}. Even when relevant information is retrieved, agents often suffer from "prior-bias," where internal model weights override the provided evidence, leading to incorrect synthesis \cite{lewis2020retrieval}. Furthermore, in dynamic environments, the temporal decay of information becomes a critical factor; agents often fail to refresh their internal state when search engine results contradict their training data \cite{vu2024freshllms}. This failure to properly integrate retrieved context often manifests as "Summarize Misalign," where the agent's internal synthesis misrepresents the very documents it successfully retrieved \cite{ji2023survey, dharmaratnakar2026generative}.

\subsection{Internal Reasoning Hallucinations}
A distinct class of failure occurs during the deliberation phase, where the agent processes retrieved information to form a logical conclusion. Even with "perfect retrieval," agents can suffer from "Factual Reasoning" errors, making incorrect logical inferences over provided context \cite{yao2022react}. This stage is also where "Math and Science Reasoning" failures emerge, as agents perform incorrect calculations or symbolic interpretations that invalidate the trajectory \cite{wei2024long}. Notably, reasoning-heavy models often fall into the "Solvability Hallucination" trap, misjudging task complexity and believing an unsolvable task is solvable due to an over-reliance on internal logic over environment constraints \cite{huang2026mm}.

\subsection{Execution and API Synthesis}
The most distinctive failure mode in industrial agentic IR occurs at the execution interface. This is where linguistic fluency and functional correctness diverge most sharply. Agents frequently exhibit "Tool Fabrication," invoking non-existent APIs or imagining capabilities that are absent from the provided SDK \cite{qin2024toolllm}. Even when the correct tool is identified, agents often specify "Incorrect Arguments," generating hallucinated parameters or invalid JSON schemas that cause system-level crashes \cite{patil2024gorilla,vijayakumar2016dsl}. These execution-level errors are often masked by the model’s reasoning trace, creating a "Fluency Trap" where the agent provides a perfectly worded justification for an action that is technically impossible to perform \cite{schick2023toolformer}.

We categorize these failures across four functional stages as detailed in Table 1.

\section{The Operational Gap: The Fluency Trap}

The transition from research benchmarks to production environments reveals a critical operational gap: the disconnect between linguistic coherence and functional grounding. We identify this as the "Fluency Trap," a phenomenon where the model’s drive for helpfulness overrides its technical accuracy.

\subsection{Functional Misalignment and the Likelihood Trap}
In industrial IR, models are often optimized via reinforcement learning to be as helpful and conversational as possible \cite{ouyang2022instructgpt}. However, this alignment can be counterproductive in agentic workflows. When an agent encounters an error—such as a failed API call or a missing tool—it often falls into a "likelihood trap" \cite{yin2025reasoning}. Instead of reporting the system error, the agent generates a plausible-looking reasoning trace that "fills in the gaps" to maintain the conversation's flow. This ensures the response remains linguistically fluent while becoming functionally disconnected from the environment's state \cite{bender2021dangers}.

\subsection{The Snowball Effect in Long-Horizon Tasks}
As trajectories increase in length, the reliability of the system decays exponentially, a phenomenon analogous to compounding error in sequential decision-making under distribution shift~\cite{ross2011dagger,bengio2015scheduled}. In long-horizon agentic search, a minor retrieval error at the start of a ten-turn trajectory can make the final result entirely irrelevant, even if every individual step appears logical in isolation \cite{gao2025beyond}. This "Snowball Effect" is particularly dangerous because current monitoring tools often focus on step-wise token probability rather than the integrity of the causal chain. In production, this leads to "Autonomy Abuse," where an agent executes a sequence of harmful or costly actions based on an initial hallucinated premise that was never verified.

\subsection{Representation Collapse in Tool Judgment}
Preliminary industrial observations suggest a 'Reasoning Trap' \cite{yin2025reasoning}, where models optimized for long-form deliberation via RL show a propensity to prioritize linguistic rationalization over functional grounding. We hypothesize this stems from a representation shift where solvability detection is suppressed in favor of completion. \cite{nanda2023mechanistic}. This results in an agent that is "confidently wrong," prioritizing a rationalized explanation over a factual grounding in the available SDK constraints~\cite{lin2022truthfulqa,manakul2023selfcheckgpt}.

\section{The Industry Position: Toward Trajectory Integrity}

Our position is that the industry must transition from measuring global output accuracy to a focus on trajectory integrity and causal attribution. For autonomous agents to be safely deployed in production, we must prioritize the "correctness of the process" over the "plausibility of the result."\cite{xu2024hallucination}

\subsection{From Output Checking to Verification Gating}
We advocate for the implementation of Verification Gates at every interaction unit $(c, a, o)$. We model each interaction step as a triple $(c_t, a_t, o_t)$, where $c_t$ denotes the internal reasoning state, $a_t$ the externally executed action, and $o_t$ the resulting environment observation, forming a trajectory over which reliability can be evaluated.This perspective aligns with selective prediction frameworks, where models abstain under calibrated uncertainty thresholds to manage risk--coverage trade-offs~\cite{geifman2017selective,guo2017calibration,lakshminarayanan2017deepensembles}. This proposal aligns with uncertainty propagation frameworks such as UProp, which decomposes per-step uncertainty into intrinsic and extrinsic components to identify optimal intervention points. To transform this into an actionable industrial standard, we propose the following stage-specific design patterns:

\begin{itemize}
\item \textbf{Planning Gates (Solvability Classifiers):} At the goal-decomposition stage, the system must check if the generated thought $c_t$ is executable within the current SDK and budget constraints. If a sub-goal is identified as unsolvable, the policy should trigger a systematic abstention.

\item \textbf{Reasoning Gates (Stepwise Attribution):} Drawing on the principles of SPA-RL\cite{wang2025spa}, these gates assign progress signals to individual reasoning steps. If the progress attribution falls below a threshold, the system must discard the current $c_t$ and re-sample the reasoning trace.

\item \textbf{Execution Gates (State-Diff \& Provenance):} Before committing to an action $a_t$, the system should utilize state externalization frameworks like InfiAgent\cite{yu2026infiagent} to record a deterministic provenance of the environment state. This allows for "dry-run" simulations where the predicted change is verified against safety protocols before the live API call is executed.
\end{itemize}

\subsection{Causal Attribution and Observability}
To manage the operational risks of agentic loops, industrial systems require granular observability that transcends simple pass/fail outcomes. This requirement parallels evaluation in interactive IR, where effectiveness must be measured over entire sessions rather than single ranked outputs~\cite{kelly2009interactiveir,smucker2012timebiased,zhang2017formalframework}. We argue for Causal Attribution to pinpoint the specific "responsible step" of a failure—a necessity for standardizing Trust, Risk, and Security Management (TRISM) \cite{raza2025trism}. We propose that practitioners adopt the following quantitative metrics to measure trajectory reliability:
\begin{itemize}
\item \textbf{First-Error Position (FEP):} This metric identifies the specific index $t$ in a trajectory of length $N$ where the agent's state first diverges from the ground truth or functional constraints. Lower FEP values indicate fundamental planning or retrieval vulnerabilities, while high FEP values suggest late-stage cumulative drift.
\item \textbf{Abstention Precision and Recall:} Given the risks of the "Fluency Trap," agents must be evaluated on their ability to correctly identify unsolvable tasks \cite{zhang2024toolbehonest}. Precision measures how often an agent's refusal is technically justified, while recall measures how many unsolvable "likelihood traps" were successfully intercepted.
\item \textbf{Rollback Recovery Rate (RRR):} This measures the percentage of trajectories where a \textbf{Verification Gate} successfully intercepted an error and the system was able to recover via a rollback policy or query refresh. This is a critical KPI for measuring the ROI of implementing per-step oversight.
\item \textbf{Weakest-Link Reliability:} Instead of averaging confidence, this score reflects the minimum probability across all $(c, a, o)$ units in a trajectory. This highlights "brittleness" in long-horizon tasks where a single low-confidence reasoning step invalidates the entire mission.
\end{itemize}

By adopting evaluation suites like OdysseyArena\cite{xu2026odysseyarena}, which specifically probe for long-horizon inductive failures, organizations can move toward an auditable, stateful lifecycle. This observability must include state externalization, leveraging frameworks like InfiAgent\cite{yu2026infiagent} to provide deterministic provenance for every tool call.

\subsection{Prioritizing Systematic Abstention}
Finally, we contend that the industry must rebalance the trade-off between "helpfulness" and "honesty". The Reasoning Trap\cite{yin2025reasoning} highlights a critical risk: that RL optimized purely for task reward can incentivize the masking of functional failures with linguistic fluency. To mitigate this, we advocate for the following industrial practices:

\begin{itemize}
\item \textbf{Cost-Sensitive Abstention Gates:} Practitioners should adopt frameworks like \textbf{PRISM}, which implement gates that only trigger an action if the model's confidence exceeds a specific risk/cost threshold. This prevents "Autonomy Abuse" by ensuring that high-risk tool calls are gated by calibrated probabilities.
\item \textbf{Process-Level Uncertainty Propagation (UQ):} Using step-level uncertainty decomposition to separate intrinsic model uncertainty from environment-induced uncertainty. This allows the system to trigger "selective reflection" or retrieval refreshes before an error snowballs into a hallucinated result.
\item \textbf{Reliability-Aligned Reward Design:} Instead of standard preference-based rewards, we propose incorporating grounding signals as a primary success metric. Drawing on SPA-RL (Stepwise Progress Attribution) \cite{wang2025spa}, industry practitioners should redistribute RL credit to reward agents for systematic abstention on unsolvable tasks, effectively training for "solvability detection".
\item \textbf{Trial-Before-Execution (Dry-Running):} Convergent with our execution gates, performing schema validation or sandboxed simulation of an API call prior to commitment reduces execution-level hallucinations and provides a deterministic anchor for action correctness.
\end{itemize}

\section{Conclusion \& Future Outlook}

The transition of Information Retrieval from document ranking to autonomous agency introduces systemic risks that conventional metrics fail to capture. The Fluency Trap and Reasoning Trap are not merely engineering issues but structural misalignments in how agentic systems are optimized and evaluated\cite{yin2025reasoning}. As compounding error remains inherent in long-horizon computation\cite{xu2024hallucination}, industrial focus must shift from model-centric scaling toward architecture-centric reliability.

We argue for a trajectory-level perspective grounded in verification gates, causal attribution, and calibrated abstention. Reliable Agentic IR systems must prioritize process correctness and grounded execution over plausible completion. Establishing trajectory integrity as a first-class evaluation objective is essential for deploying production-ready autonomous systems.

\section*{Acknowledgements}
The authors acknowledge the use of Google’s Gemini AI
to refine the text and generate the images included in this publication.

\section{Presenter Biography}
Anushree Sinha is a Software Engineer and AI Researcher at Google with 12 years of professional experience, specializing in the reliability and factuality of Large Language Models (LLMs) for the past three years. She currently leads technical GenAI initiatives at Google, where she spearheaded the architecture of novel automated measurement frameworks to reduce hallucination rates in high-traffic production systems. Her recent work includes designing robust multilingual grounding pipelines and developing end-to-end infrastructure for generative search features. Prior to her focus on LLMs, she built the "Actions SDK" for Google Assistant and managed real-time bidding infrastructure at Yelp. Anushree is a lead author on active research submissions regarding automated grounding evaluation and holds an M.S. in Computer Science (Machine Learning) from the Georgia Institute of Technology. Her work focuses on bridging the gap between theoretical AI safety and the practical demands of production environments.

\bibliographystyle{ACM-Reference-Format}
\bibliography{references}
\end{document}